\documentclass[11pt]{article}

\usepackage[final]{acl}
\usepackage{multirow}
\usepackage{times}
\usepackage{latexsym}
\usepackage{enumitem}
\usepackage{subcaption}
\usepackage{todonotes}
\usepackage{booktabs}
\usepackage[T1]{fontenc}
\usepackage{hyperref}
\usepackage[utf8]{inputenc}

\usepackage{microtype}

\usepackage{inconsolata}

\usepackage{graphicx}
\usepackage{amsmath}
\usepackage{dsfont}
\title{From Vision to Language: Investigating Causal Information Flow in Multimodal Decision-Making}

\author{
    Davide Testa\textsuperscript{1,2}, Hugh Mee Wong\textsuperscript{3},
    \textbf{Alessandro Lenci\textsuperscript{4}}, \textbf{Bernardo Magnini\textsuperscript{1}}, \textbf{Albert Gatt\textsuperscript{3}} \\
    \small \textsuperscript{1}Fondazione Bruno Kessler (FBK) - Trento, Italy, \textsuperscript{2}Università di Roma La Sapienza - Rome, Italy, \\ 
    \small \textsuperscript{3}Utrecht University - Utrecht, The Netherlands,
    \small \textsuperscript{4} University of Pisa - Pisa, Italy,\\
    \small \texttt{\{dtesta, magnini\}@fbk.eu}, 
    \small \texttt{ alessandro.lenci@unipi.it}, \small \texttt{\{h.m.wong, a.gatt\}@uu.nl}
}

\begin{document}
\maketitle
\begin{abstract}
Vision-Language Models are commonly evaluated through their final predictions, but understanding whether these decisions are grounded in visual evidence requires tracing how visual information contributes to language-based decisions. With this in mind, we investigate cross-modal information flow in a video-based generative multiple-choice-like setting by applying causal interventions to video-text attention pathways and examining its effects across layers.
We target spatial, causal, and temporal visual reasoning. Our results show that visual information is integrated mainly when the model processes the candidate answer options, which serve as the primary  textual grounding sites for the final decision. 
We further show that nouns play an important role as semantic anchors during multimodal enrichment, while verbs are more relevant when temporal relations are processed.
Finally, we identify a distinct pattern in temporal reasoning, suggesting that VLMs struggle to reconstruct sequential information across video frames, but we remark that such fragility may also reflect linguistic biases associated with specific temporal expressions used for defining the relation between events within a scene.\\ 
Code and data available on \href{https://github.com/Caput97/Multimodal_Information_Flow_on_video-Decision-Making.git}{Github}.

\end{abstract}

\section{Introduction}
\label{sec:intro}

\begin{figure}[t]
  \centering
    \includegraphics[width=\linewidth]{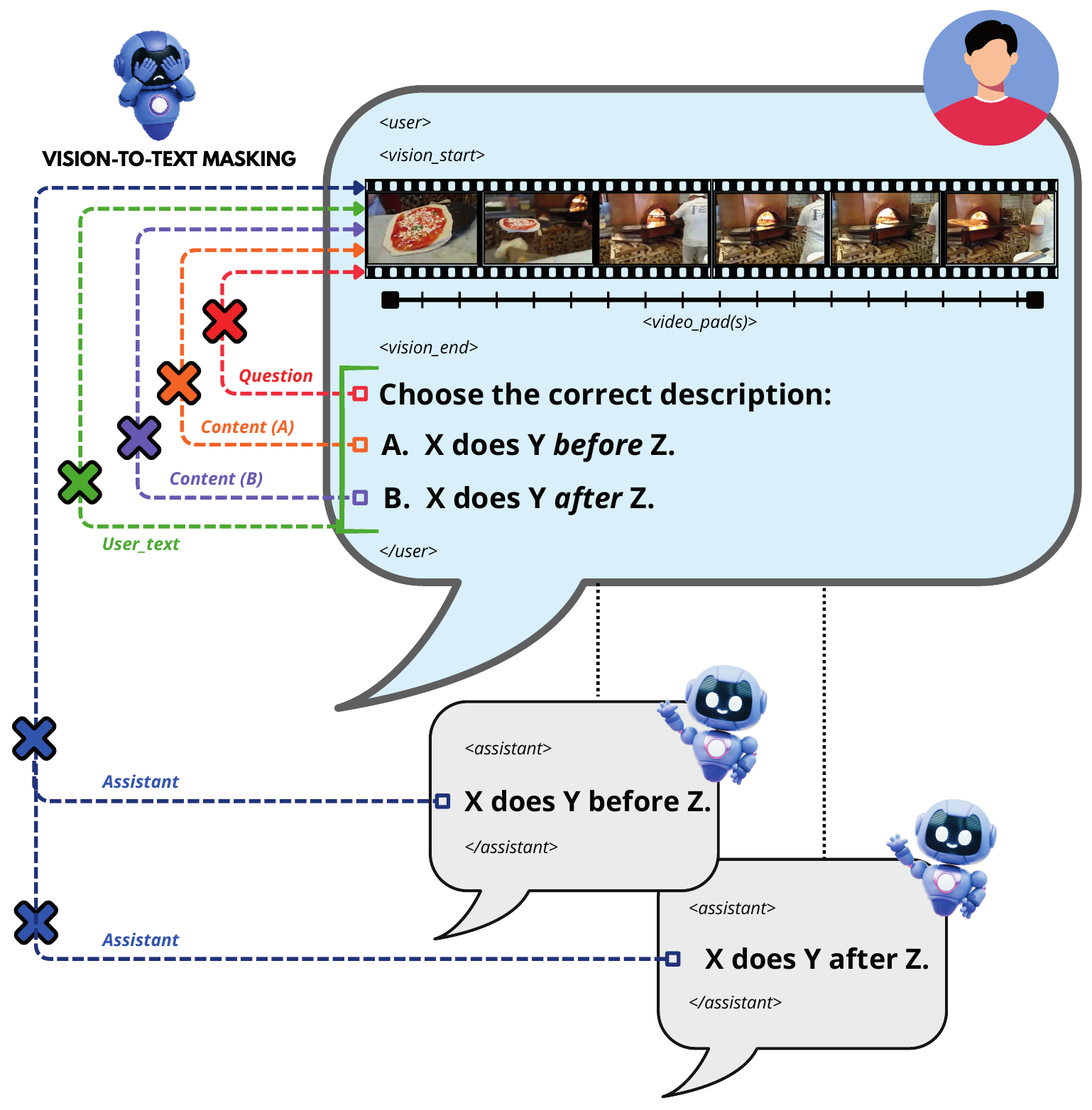}
    \label{fig:prompt_VT}
  \hfill

  \caption{Overview of the Attention Knockout approach with prompt structure and region-based masking.}
  \label{fig:prompt}
\end{figure}


Modern Large Language Models (LLMs) and Vision-Language Models (VLMs) are commonly evaluated through benchmark-based and behavioral assessments. In this setting, abilities such as language understanding \cite{wang-etal-2018-glue, NEURIPS2019_4496bf24}, visual grounding \cite{parcalabescu-etal-2022-valse}, or event reasoning \cite{mvbench} are inferred from the model's input-output behavior on diagnostic tasks \cite{warstadt-etal-2020-blimp-benchmark, ribeiro-etal-2020-beyond, chang-bergen-2024-language}. While this provides an essential way to compare models and quantify their performance, it does not explain which internal computations support a given prediction, nor whether models are relying on their input in the expected manner.
This limitation is especially relevant for VLMs: in multimodal settings, a correct answer may suggest that the model has successfully integrated visual and linguistic information, but the final prediction alone does not reveal how this integration is implemented internally. The model may rely on visual evidence, exploit regularities in the textual prompt, or combine both in ways that remain difficult to disentangle \cite{wu2026visionlanguagelayerwiseinformationtheoretic}. As a result, evaluating whether a VLM can solve a task is not sufficient to understand how visual information contributes to its decision \cite{sim-etal-2025-vlms}. This problem becomes even more pronounced when videos are provided as input. Compared to static images, videos require models to process visual information across time, including objects, actions, event dynamics, and relations such as spatial configurations, causality, and temporal order \cite{mvbench, liu-etal-2024-tempcompass, loginova-loguinova-2025-deep, song-etal-2025-burn}. These dimensions make video-based tasks a useful testbed for studying multimodal reasoning, as they require us to trace how visual information flows through the model's computation and supports language-based decisions. \\
To understand how visual evidence supports multimodal decisions, this paper investigates the internal cross-modal dynamics through which VLMs use video information in a decision-making setting. We focus on two research questions: (\textbf{RQ1}) Where and how does visual information interact with the textual representations used by the model to choose between alternative video descriptions? (\textbf{RQ2}) How does cross-modal information contribute across spatial, causal, and temporal reasoning?

We address the questions above in a controlled multimodal decision-making task in Italian, where a model must select the correct description between a caption and a minimally different foil associated with the same video. To study this process mechanistically, we apply Attention Knockout to video-text attention pathways and measure how blocking specific cross-modal interactions affects the plausibility of candidate answers.\\
\textbf{Contributions.}  (i) a causal intervention framework for tracing \textit{video-to-text} information flow in VLMs during caption--foil decision-making; (ii) an analysis of the prompt regions and token types that mediate visual grounding; (iii) an investigation of how these dynamics vary across spatial, causal and temporal relations.

The rest of this paper is structured as follows. Section~\ref{sec:relWorks} reviews the related work and contextualizes our approach. Section~\ref{sec:exp} describes the experimental setup, including the data (\ref{subsec:data}), models (\ref{subsec:models}), and methodology (\ref{subsec:methodology}). Section~\ref{sec:resDisc} presents and discusses the results. Finally, Section~\ref{sec:concl} summarizes the main findings and concludes the paper.

\section{Related Work}
\label{sec:relWorks}


In vision-language research, one influential line of behavioral evaluation is represented by \textbf{foiling methods} 
\cite{shekhar-etal-2017-vision, shekhar-etal-2017-foil}, 
which require the model to discriminate between alternatives that differ only in a targeted semantic element, turning visual understanding into a fine-grained contrastive problem \cite{madhyastha-etal-2018-defoiling, parcalabescu-etal-2022-valse, kesen2023vilma, samin-etal-2025-colorfoil}. 
However, these methods do not reveal how visual information is internally processed and used to support model decisions. Indeed, a correct answer does not necessarily imply that the model has grounded its decision in the relevant visual evidence, since it may instead exploit linguistic priors or dataset regularities, reaching the correct choice by relying only weakly, or not at all, on the visual modality. This type of failure, often referred to as \textit{unimodal collapse}, highlights the need for analyses that go beyond final task accuracy \cite{parcalabescu-etal-2021-seeing, parcalabescu-frank-2023-mm, chen-etal-2024-quantifying, sim-etal-2025-vlms}. \\ This motivates a shift from behavioral studies to \textbf{Mechanistic Interpretability} (MI) research, which aims to identify the representations, circuits, and information flows that causally support model predictions.
In unimodal language models, this line of work has produced influential accounts of transformer computation, including circuits for in-context learning and induction heads \citep{elhage2021-circuits, olsson2022-induction}, factual recall and knowledge localization \citep{meng2022-edit, geva-etal-2023-dissecting}, and multi-hop or arithmetic reasoning \citep{stolfo2023-arithmetic, biran2024-hopping}.
More recently, these methods have been extended to VLMs, where they have been used to study how visual information is localized and transformed across layers. These analyses suggest that visual representations can evolve toward more text-like forms and that visual information flow varies across reasoning tasks \citep{ICLR2025_900fb343, zhang-etal-2025-redundancy}.

Attention mechanisms are an important target for mechanistic investigation as they regulate how information is exchanged across tokens and composed into internal representations.
However, attention patterns alone are not sufficient for causal explanations, motivating intervention-based methods that operate directly on attention connections.
Recent work has analyzed attention heads and attention-based information flow in VLMs, identifying heads involved in multimodal integration, grounding, object localization, and image-to-text-transfer \citep{golovanevsky2025-notice, kang2025-grounding, kim2025interpreting}.
Among these methods, \textbf{Attention Knockout} selectively blocks attention edges between subsets of tokens, effectively performing activation-level ablations on attention pathways \cite{geva-etal-2023-dissecting}.
This enables researchers to block attention from image tokens to query or generated tokens and test where visual information is used during inference \citep{ICLR2025_900fb343, kaduri2025-image, zhang-etal-2025-redundancy}.
However, most existing work on VLMs still focuses on image-based tasks \cite{ortu-etal-2024-competition, Wang_2025_CVPR, Zhang_2025_CVPR, ICLR2025_900fb343, cohen-etal-2025-performance}, leaving open questions about how modality-specific circuits emerge and interact in more complex multimodal scenarios such as videos.

Our work connects two previous lines of research by applying attention knockout to a caption-foil discrimination problem.
This design combines the structure of multiple-choice question-answering evaluation with the fine-grained control afforded by foiling methods,
while also allowing us to examine how the model internally supports its decision.
\section{Experimental Setting}
\label{sec:exp}

\begin{table*}[t]
  \centering
  \small
  \begin{tabular}{lll}
    \hline
    \textbf{Dataset} & \textbf{Option type} & \textbf{Sentence} \\
    \hline

    \multirow{2}{*}{\textsc{Spatial}}
      & \textit{Caption} & At the end of the scene, the little girl
walks \textbf{on a carpet}. \\
      & \textit{Foil}    & At the end of the scene, the little girl walks \textbf{on a path}. \\

    \multirow{2}{*}{\textsc{Temporal}}
      & \textit{Caption} & \textbf{Right after} cutting the pizza, the girl hands it to a person in front of her. \\
      & \textit{Foil}    & \textbf{Before} cutting the pizza, the girl hands it to a person in front of her. \\

    \multirow{2}{*}{\textsc{Causal}}
      & \textit{Caption} & The girl is upset because \textbf{of a cockroach}. \\
      & \textit{Foil}    & The girl is upset because \textbf{of a loud noise}. \\

    \hline
  \end{tabular}
  \caption{
  Caption--foil examples from MAIA subsets (English translations; originals in Table~\ref{tab:maia_examples_ita}, Appendix~\ref{app:add_data}).}
  \label{tab:maia_examples}
\end{table*}

We consider a generative caption--foil task based on video understanding. Given a multimodal prompt pairing a video with a text (Figure~\ref{fig:prompt}), the model is asked to select the most plausible description among two candidate options
. The candidates correspond to a \textit{caption} and a \textit{foil} linked to the video, where the foil is a minimally altered version of the caption designed to introduce a subtle but semantically critical inconsistency \cite{shekhar-etal-2017-foil}. This setup enables fine-grained evaluation of the model’s ability to ground linguistic reasoning in visual content.

\subsection{Data}
\label{subsec:data}
To construct our evaluation data, we leverage subsets of the MAIA benchmark \cite{testa-etal-2025-one, testa-etal-2025-maia}, which provides carefully curated Italian \textit{caption–foil} pairs associated with videos. 
In particular, we focus on three subsets capturing different types of reasoning: (i) \textsc{\textbf{Spatial}}, involving spatial relations between entities; (ii) \textsc{\textbf{Temporal}}, involving ordering and dynamics of events; (iii) \textsc{\textbf{Causal}}, involving cause-effect relationships.
Each subset comprises $1{,}600$ caption--foil pairs associated with the $100$ videos in the dataset, corresponding to $16$ pairs per video within each subset and $4{,}800$ pairs overall.
Table~\ref{tab:maia_examples} shows representative caption--foil pairs from the three MAIA subsets. The assignment of the caption and foil to option letters \textit{A} / \textit{B} within the decision-making task is then randomized to avoid position biases.

By controlling the semantic and lexical differences within caption--foil pairs, and by considering subsets that capture distinct types of reasoning, this setup provides a controlled testbed to probe and disentangle the contribution of visual information to model predictions both at the instance level and across different reasoning categories.

\subsection{Models}
\label{subsec:models}
We tested \textbf{Qwen2.5-VL 7B}~\cite{qwen2.5technicalreport} and \textbf{LLaVA-OneVision 7B}~\cite{llavanextvideo, li2024llava1V}, two VLMs from different model families that both pair a vision encoder with a decoder-only language model.
In our experimental setting, each video is sampled into $16$ frames, whose patches are processed by a vision encoder and mapped into visual token embeddings. These embeddings are concatenated with the textual tokens and jointly processed through self-attention within the language-model backbone. This shared architecture allows us to apply the same experimental framework to both models.

Despite this shared paradigm, the models differ in their visual and spatiotemporal encoding, positional representations, and training design, allowing us to assess whether these choices affect visual information flow.\footnote{Both models are available on the Hugging Face Hub. For further information, see Appendix \ref{sec:appendixB}.}

\subsection{Methodology}
\label{subsec:methodology}
To analyze how visual information contributes to model predictions, we adopt the established \textbf{Attention Knockout} methodology. Building on work by \citet{Zhang_2025_CVPR}, our implementation intervenes directly on the pre-softmax attention-score matrices during the forward pass, rather than relying on hook-based approximations. 
The key idea is to selectively block specific attention connections and measure the resulting effect on model outputs. Concretely, we intervene directly on the self-attention mechanism of the VLM language backbone by applying a patch function (i.e., a targeted modification) to the original attention forward pass, without altering its underlying causal nature. Given a sequence of tokens, the classical attention operation $\mathrm{Attention}(Q,K,V) = \mathrm{softmax}\left(\frac{QK^{\top}}{\sqrt{d_k}}\right)V$ can be interpreted as a set of interactions (i.e., the Attention matrix) between query tokens and key tokens, where each query token attends to key tokens in the sequence to gather relevant contextual information. Our framework selectively removes subsets of these interactions by modifying the attention mask during the forward pass.
For each transformer layer, we define a set of masking constraints over attention connections in the form of tuples $(q, k_{start}, k_{end})$, where $q$ is a query token index and $(k_{start}, k_{end})$ defines a contiguous span of key tokens. For each of these tuples, attention from query $q$ to all keys in the specified span is blocked by setting the corresponding attention scores to $-\infty$, effectively removing their contribution after the softmax computation.
The same masking constraints are applied globally across all transformer layers in a single forward pass, with no layer-wise sliding window, 
enabling us to study how information flows through the network under controlled perturbations. 
A key aspect 
is the explicit separation between query tokens, which define where the intervention is applied, and key spans, which define the source of information being suppressed. This separation allows us to model different directions of information flow.
In the main analysis, we focus on the \textit{vision-to-text} direction, blocking attention from textual queries to visual tokens to directly test whether textual tokens in the prompt acquire task-relevant information from the visual stream and integrate visual evidence into language-token representations. The complementary \textit{text-to-vision} direction blocks attention from visual queries to textual tokens, and works as an additional control analysis (see Appendix~\ref{app:text-to-vision}).
This intervention design enables fine-grained control over cross-modal interactions and provides a way to probe the contribution of specific input components to the model’s internal computations.

\textbf{Region-Based Attention Knockout.}
We build the attention knockout framework in our multimodal task setting by defining semantically meaningful regions over the input sequence (see Figure~\ref{fig:prompt}).
Given the structured prompt, we identified the following regions: the \textit{question}\footnote{In our setting, the “question” denotes the task instruction provided to the model (i.e., \textit{Choose the correct description}), rather than an interrogative sentence about the video content.}, the content of option \textit{A}, the content of option \textit{B}, the \textit{assistant content} (i.e., all the response tokens), and a \textit{vision} region corresponding to the tokens encoding the video input. These regions are automatically identified through token-level position mappings.\\
In the main analysis, we apply region-based masking in the \textit{vision-to-text} direction by selectively blocking attention from each of these textual regions to the vision one. 
This allows us to trace the flow of information from video to the prompt components and generated output.\footnote{For completeness, we also evaluated the complementary \textit{text-to-vision} direction in Appendix \ref{app:text-to-vision}, which requires an inverted text-video prompt under causal attention.}
 

\textbf{Token-level Attention knockout.}
While region-based masking enables a coarse-grained analysis of cross-modal interactions, it does not capture which specific linguistic elements within a region are responsible for the observed effects. We therefore extend our framework to a finer-grained token-level masking strategy, where attention constraints are defined over subsets of tokens. In this experimental setup, we focus on the content of options A and B within the multiple-choice prompt, as they represent the central decision points of the task.
Concretely, instead of treating A- and B-content regions as homogeneous spans, we decompose them into subsets of tokens based on their linguistic properties. In particular, we leverage part-of-speech (POS) annotations\footnote{PoS parsing done with the spaCy \textit{it\_core\_news\_sm} model.} to isolate semantically relevant categories, including \textsc{nouns}, \textsc{verbs}, and \textsc{prepositions}, as well as a subset-specific group, \textsc{category-critical tokens}, corresponding to the minimal lexical elements that distinguish the caption from its foil and determine the reasoning category targeted by the pair. For example, consider the following pair in the {\em causal} subset: \textit{The man fell into the water because he \underline{lost his balance}} versus \textit{The man fell into the water because he \underline{was pushed}}. The underlined spans correspond to the category-critical tokens, as they encode the alternative causal explanations that determine which candidate is correct with respect to the video input.
We applied the same $(q, k_{start}, k_{end})$ formulation for deleting specific attention edges and extended it to token-level granularity by operating on these subsets of tokens. For instance, under the \textit{vision-to-text} direction, we can block the contribution of visual tokens when attending to noun tokens within option A, thereby isolating the role of object-level grounding. Similarly, by masking verb or preposition tokens, we can probe how action and relational cues contribute to the integration of visual information.\\
Compared to region-based masking, this token-level approach provides a more precise characterization of the information flow within the model, enabling us to disentangle the contribution of different linguistic features to multimodal reasoning.\\[4pt]
For both region- and token-based masking, we obtain layer-wise scores from the same globally masked forward pass using logit-lens readout \cite{nostalgebraist2020logitlens}.
At each layer, we apply the model's final normalization and language-model head to the intermediate representations. We then use a \textit{forced-target scoring} procedure: each caption or foil is inserted in the assistant-response position, as shown in Figure~\ref{fig:prompt}, and the probability of each token is computed autoregressively, conditioned on the multimodal input and the preceding target tokens. The sentence-level log-probability is then obtained by summing the autoregressive log-probabilities of its tokens, allowing us to compare the plausibility assigned to the two candidate sentences.
%
%

\begin{table}[t]
  \centering
  \scriptsize
  \begin{tabular}{llcc}
    \hline
    \textbf{Model} & \textbf{Dataset} & 
    \textbf{Acc$_{\text{ A / B}}$} &
    \textbf{Acc$_{\text{ caption / foil}}$} \\
    \hline

    \multirow{3}{*}{Qwen2.5-VL}
      & \textsc{Spatial}   &  0.69 & 0.67  \\
      & \textsc{Temporal}  & 0.77 & 0.43  \\
      & \textsc{Causal}    & 0.84 & 0.81  \\
    \hline

    \multirow{3}{*}{LLaVA-OneVision}
      & \textsc{Spatial}   & 0.69 & 0.68  \\
      & \textsc{Temporal}  & 0.73 & 0.44  \\
      & \textsc{Causal}    & 0.83 & 0.81  \\

    \hline
  \end{tabular}
  \caption{Behavioral performance on the caption–foil task for the three MAIA subsets, under two scoring schemes: A/B labels vs. full candidate sentences.}
  \label{tab:behavioral_accuracy}
\end{table}

As a preliminary check (Table~\ref{tab:behavioral_accuracy}), we compare full-sentence scoring with A/B-label scoring,\footnote{For A/B label scoring, we append \textit{Answer only with A or B} to the original prompt in Figure~\ref{fig:prompt}, thereby replicating the original experimental setting. 
For full-sentence scoring, the predicted answer is the candidate with the higher forced-target sentence score; accuracy is therefore computed as
$\frac{1}{N}\sum_{i=1}^{N}\mathds{1}\left[
\log P_M(C_i \mid x_i) >
\log P_M(F_i \mid x_i)
\right]$,
where $C_i$ and $F_i$ denote the caption and foil, respectively; $x_i$ is the corresponding multimodal input; and $M$ is the model.} which reproduces the behavioral trends reported in \citet{testa-etal-2025-one}. 
Full-sentence scoring yields comparable results on Spatial and Causal items but substantially lower accuracy on Temporal items, with performance dropping below chance level.\footnote{We nevertheless adopt full-sentence scoring because it provides the token-level plausibility scores required for our mechanistic analyses.} Comparison with A/B-label scoring suggests that this drop may partly reflect lexical biases in autoregressive sentence scoring, particularly around temporal markers, rather than temporal reasoning alone (see Section~\ref{subsec:temporality_issue}).

\section{Results}
\label{sec:resDisc}

\begin{figure*}[t]
  \centering
    \includegraphics[width=\linewidth]{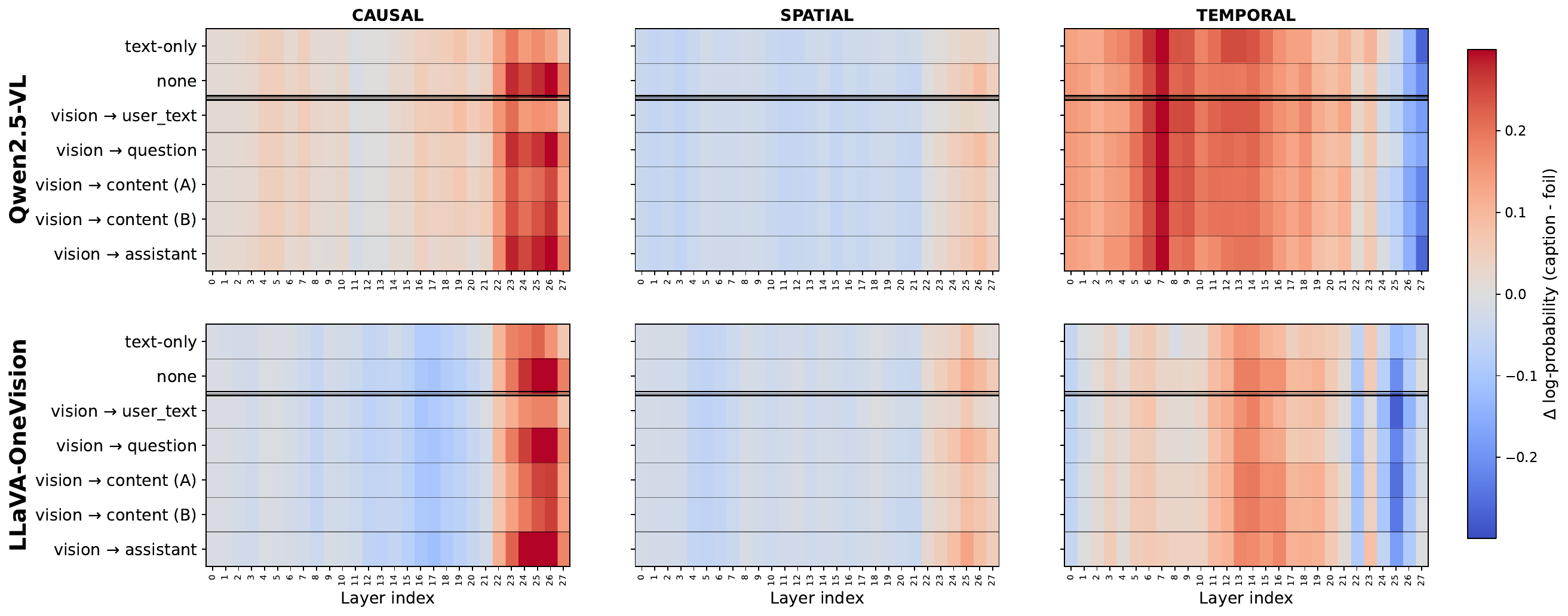}

  \caption{Region-based attention knockout results. Rows indicate the textual prompt region prevented from attending to visual tokens (i.e., \textit{Vision} $\rightarrow$ \textit{Region}). Colors show the layer-wise caption--foil log-probability difference. \textit{Text-only} and \textit{None} condition represent our two baselines.}
  \label{fig:heatmaps}
\end{figure*}

\begin{table*}[htbp]
\centering
\resizebox{\textwidth}{!}{%
\begin{tabular}{llrr|*{5}{rr}}
\toprule
\multirow{2}{*}{Model} & \multirow{2}{*}{Dataset} &
\multicolumn{2}{c|}{text-only Vs. vision} &
\multicolumn{2}{c}{vision$\to$question} &
\multicolumn{2}{c}{vision$\to$content(A)} &
\multicolumn{2}{c}{vision$\to$content(B)} &
\multicolumn{2}{c}{vision$\to$user\_text} &
\multicolumn{2}{c}{vision$\to$assistant} \\
\cmidrule(lr){3-4}
\cmidrule(lr){5-6}
\cmidrule(lr){7-8}
\cmidrule(lr){9-10}
\cmidrule(lr){11-12}
\cmidrule(lr){13-14}
& & Acc & Acc
  & Flip rate & $\Delta$Acc
  & Flip rate & $\Delta$Acc
  & Flip rate & $\Delta$Acc
  & Flip rate  & $\Delta$Acc
  & Flip rate & $\Delta$Acc \\
\midrule
\multirow{3}{*}{Qwen2.5-VL}
 & \textsc{Spatial}  & 0.54 & 0.67 & 8.19\%  & $+0.06$ & 17.62\% & $-5.0$ & 15.0\% & $-2.62$ & \textbf{26.38\%} & $-11.75$ & 8.38\%  & $-0.37$ \\
 & \textsc{Temporal} & 0.39 & 0.43 & 8.25\%  & $-1.50$ & 14.62\% & $-6.5$  & 12.12\% & $-4.5$ & \textbf{18.75\%} & $-5.13$ & 6.06\% & $-0.69$ \\
 & \textsc{Causal}   & 0.62  & 0.81 & 4.69\%  & $-0.81$ & 14.62\% & $-7.87$  & 10.88\% & $-4.87$ & \textbf{21.62\%} & $-15.50$ & 4.38\%  & $+0.25$ \\
\midrule
\multirow{3}{*}{LLaVA-OneVision}
 & \textsc{Spatial}  & 0.55 & 0.68 & 8.06\%  & $-0.56$ & \textbf{24.62\%} & $-7.87$ & 16.50\% & $-6.50$ & 23.75\% & $-10.63$ & 4.5\% & $-1.38$ \\
 & \textsc{Temporal} &  0.38 & 0.44 & 9.25\% & $-4.87$ & 18.88\% & $-0.87$  & 10.81\% & $-0.44$ & \textbf{23.81\%} & $-10.56$ & 5.62\% & $-1.00$ \\
 & \textsc{Causal}   &  0.66 & 0.81  & 5.44\%  & $-0.31$ & 19.12\% & $-8.75$  & 17.50\% & $-8.50$ & \textbf{20.06\%} & $-12.56$ & 3.19\% & $+0.19$ \\
\bottomrule
\end{tabular}%
}
\caption{Final-layer results for the \textit{text-only} baseline and region-based Attention Knockout. We report accuracy for \textit{text-only} and, for each knockout condition, Flip Rate and $\Delta$Accuracy relative to the unmasked \textit{none} condition. Flip Rate is the proportion of changed predictions ($C\to F$ or $F\to C$), while $\Delta\mathrm{Acc}=\mathrm{Acc}_{\mathrm{condition}}-\mathrm{Acc}_{\mathrm{none}}$ is the accuracy change in percentage points; negative values indicate that the intervention harms performance.}
\label{tab:flip_delta_scores}
\end{table*}

We now report the results of both the region-based and token-level attention knockout experiments.

Figure \ref{fig:heatmaps} shows, for each tested model and subsets, the layer-wise difference between the log-probability assigned to the caption and the foil under each tested masking condition, computed as $\Delta = \log P_{C} - \log P_{F}$. Positive values therefore indicate that the model assigns higher probability to the correct caption, whereas negative values indicate a preference for the foil. The results are reported across different prompt regions for which attention to the video tokens has been masked.\footnote{As noted above, we report here only the results for the \textit{vision-to-text} masking direction, since the same general trends are also observed in the \textit{text-to-vision} setting. The corresponding results are provided in Appendix \ref{app:text-to-vision}.} The \textit{none} condition corresponds to the standard multimodal forward pass without attention knockout.
The \textit{text-only} condition included in Figure~\ref{fig:heatmaps} serves as a unimodal baseline, measuring the extent to which the models can solve the task by relying exclusively on linguistic priors, without access to multimodal information from the video. Its corresponding final-layer accuracy scores, together with those of the multimodal (\textit{none}) condition, are reported in Table~\ref{tab:flip_delta_scores}.
The table further complements the layer-wise log-probability analysis with a final-layer decision-level evaluation, showing how each Attention Knockout intervention affects model predictions with reference to \textit{none}.

Across both models and all three subsets, the \textit{text-only} condition consistently underperforms the full-input \textit{none} condition, indicating that linguistic priors alone are insufficient to fully solve the task.

Having established that access to the video improves final-layer accuracy, we next examine how information is processed across layers.
Overall, both models exhibit a progressive pattern in the processing of Causal and Spatial information. This trend appears independently of the masking condition: it is visible both in the \textit{none} setting and when attention to the video is blocked for specific prompt regions. In early and middle layers, the models exhibit uncertainty with respect to the two answers, as shown by values close to zero, or even biased toward the incorrect foil, as indicated by negative values. This can be observed, for instance, around layers $4$-$6$ for the Causal subset in LLaVA-OneVision and around layer $4$ in Qwen2.5-VL. In later layers, 
from layer $22$ onward (with Qwen2.5-VL showing this shift slightly earlier), the trend reverses: both models increasingly assign higher probability to the correct caption, confirming it as the final prediction.

A different pattern emerges for Temporal information. 
In Figure~\ref{fig:heatmaps}, which averages over all items, Temporal cases appear to differ from Spatial and Causal ones, with the preference for the caption decreasing in later layers and the foil often receiving higher probability. However, the split analysis ($C>F$ vs. $F>C$) in Appendix~\ref{app:split_analysis} shows that this averaged pattern is mainly driven by the incorrect cases ($F>C$): both models 
initially favor the caption, while the foil becomes more prominent only in late layers. Thus, the apparent opposite trajectory in the averaged plot only reflects a late-layer shift in the incorrect split, indicating that Temporal caption--foil decisions are determined later in the network and are more fragile than Spatial/Causal. The difference between Causal and Spatial subsets, on the one hand, and the Temporal subset, on the other, also emerges when considering the effect of masking attention to the video across different prompt regions. For Spatial and Causal data, both models clearly show 
a different behavior with reference to the \textit{none} condition when video attention is blocked within the user-text region. This part structurally includes the question, and the two answer options A and B, as illustrated in Figure \ref{fig:prompt}. In both Spatial and Causal subsets, the strongest effect of video masking is observed when the intervention targets the two answer options, especially in late layers. Conversely, blocking the connection between video tokens and the question does not lead to substantial differences in model behavior. 
\noindent This finding is consistent with \citet{Zhang_2025_CVPR}, who showed that, in multimodal input processing, visual information is mainly used by text tokens in early and middle layers. These text tokens then propagate the visually enriched information to the assistant tokens for response generation. In other words, the visual signal survives through textual representations, which become the effective carriers of visual information, rather than the video tokens themselves. At the same time, our layer-wise analysis reveals a different temporal profile: the strongest effects of visually enriched textual tokens emerge mainly in mid-to-late layers, while direct $vision \rightarrow assistant$ masking produces only minor differences with respect to the \textit{none} condition. 
This suggests that visual information is not routed to the final decision through a direct video-to-assistant pathway, but through textual tokens that remain (visually) informative and exploited until the late stages of processing.
In the specific case of our decision-making task setting, these results suggest that cross-modal information is primarily constructed when the model processes the two candidate options. The question itself does not appear to be the critical area of multimodal grounding, and the same happens for the assistant region. Therefore, the video becomes most relevant only when the model evaluates the semantic content of the caption and foil.
The decision-level results in Table~\ref{tab:flip_delta_scores} corroborate the patterns observed in the heatmaps. Masking the answer-option content generally produces the highest flip rates and the largest changes in accuracy
. The 
negative $\Delta$Accuracy values further indicate that suppressing vision-to-option attention disrupts more correct predictions than it recovers, supporting the role of visual information during answer evaluation. Conversely, masking the question and assistant regions 
produces lower flip rates and smaller changes in accuracy, consistently with their 
limited effects in the heatmaps.

The Temporal subset again deviates from this pattern. For both models, the effect of masking video attention within the broader user text region, compared to other regions, is much less visible than in the Spatial and Causal subsets. Surprisingly, in LLaVA-OneVision we also observe an effect of the intervention around layers $20$-$21$ when masking the question region, i.e., the instruction \textit{Choose the correct description}. Moreover, this effect is more apparent in middle layers than in late layers, unlike what happens for Spatial and Causal data, where the strongest differences tend to emerge toward the end of the network. Interestingly, for the same model, at layer $23$, the $vision \rightarrow assistant$ attention knockout has a positive effect: the model assigns higher probability to the caption, in contrast both to the other masking regions and to the \textit{none} condition.
This late-layer effect is noteworthy because it occurs at a stage where multimodal information should, in principle, have already been largely processed and propagated through visually enriched text representations. This suggests that, in the Temporal subset, the model may not fully rely on such textual carriers of visual information. Therefore, it appears to still depend on direct access to the video in this final prompt region, possibly in an attempt to recover additional visual evidence needed to solve the task.\\

\begin{figure}[t]
  \centering    \includegraphics[width=\linewidth]{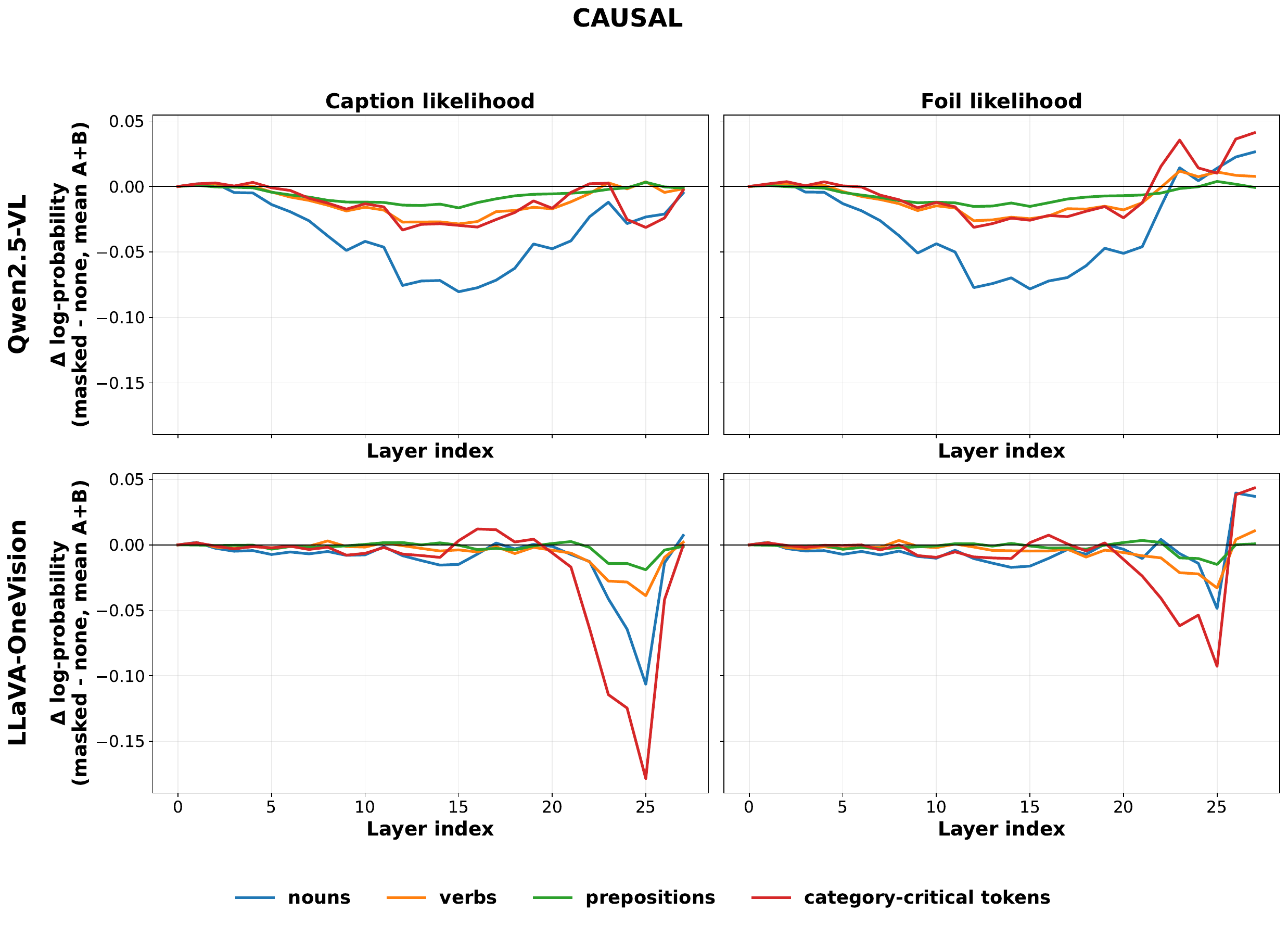}

  \caption{Layer-wise effect of token-level attention knockout on (causal) candidate log-probability.}
  \label{fig:Causal_ling}
\end{figure}

\begin{figure}[t]
  \centering
    \includegraphics[width=\linewidth]{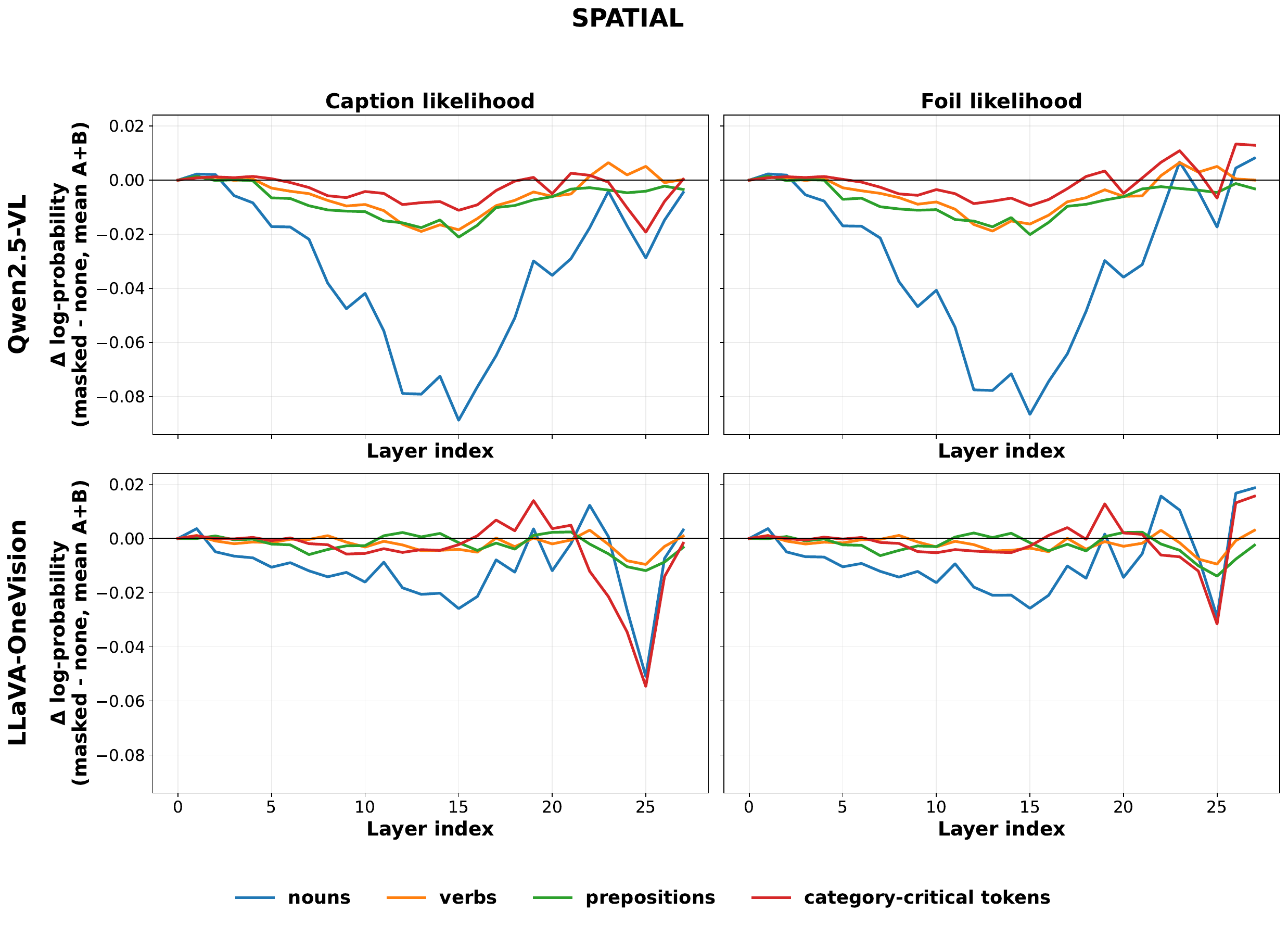}

  \caption{Layer-wise effect of token-level attention knockout on (spatial) candidate log-probability.}
  \label{fig:Spatial_ling}
\end{figure}

\begin{figure}[t]
  \centering
    \includegraphics[width=\linewidth]{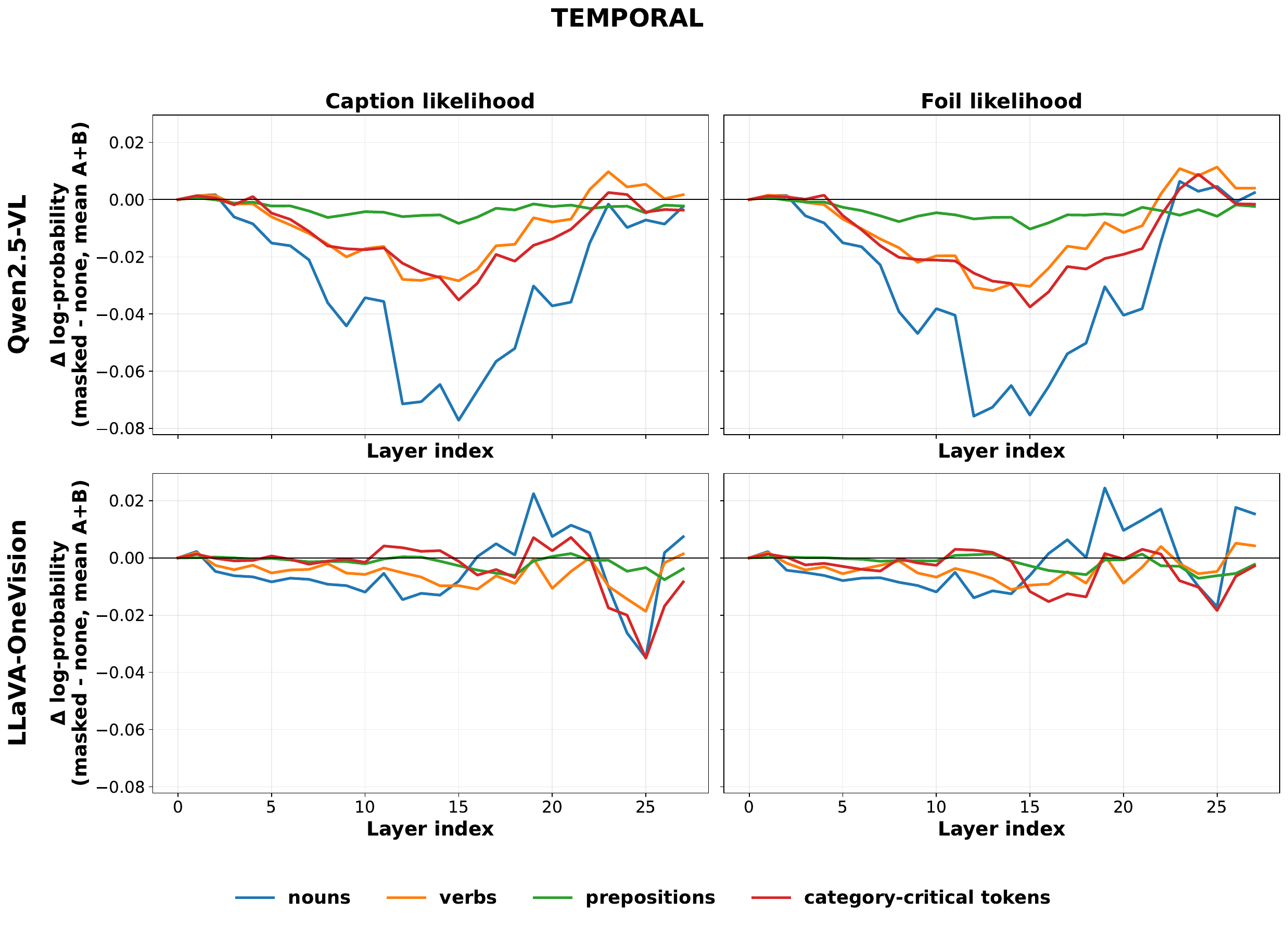}
  
  \caption{Layer-wise effect of token-level attention knockout on (temporal) candidate log-probability.}
  \label{fig:Temporal_ling}
\end{figure}

\subsection{Token-level Analysis}
Starting from 
the results above, we further investigate the part of the user-text region that emerged as the most relevant for constructing multimodal information in our setting: the two candidate options A and B. We therefore move to a finer-grained token-level analysis, asking which linguistic elements within the options are most involved in the exchange of information with the visual input.

Figures \ref{fig:Causal_ling}, \ref{fig:Spatial_ling}, and \ref{fig:Temporal_ling} show the effect of ablating the $vision \rightarrow token(s)$ connection within the A and B options. More specifically, each line shows the change in the caption or foil log-probability when \textit{vision-to-text} attention knockout is applied to specific token groups, compared to the \textit{none} condition ($\Delta = \log P_{masked} - \log P_{none}$).\footnote{For space reasons, the reported values correspond to average changes in log-probability obtained by collapsing the results across the A and B options. The same general trends are observed in the non-aggregated results.} A first clear pattern is that the strongest drop in performance across layers is observed for Qwen2.5-VL, where masking access to the video for specific linguistic elements within the options produces a more pronounced effect on both caption and foil plausibility. This suggests that Qwen2.5-VL relies more strongly on visual information while processing the candidate answers, and that grounding is actively performed in this region of the prompt. 
By contrast, LLaVA-OneVision is less affected by this intervention overall, with the exception of a consistent effect around layer $25$ across all 
subsets.

A second common trend concerns the type of linguistic elements most affected by the $vision \rightarrow token(s)$ attention knockout. Across models and subsets, nouns are consistently among the most affected categories, both when evaluating the caption and the foil. This suggests that visual information is primarily routed through lexical elements denoting visually grounded entities, which act as semantic anchors between the visual input and the linguistic representation of the candidate sentence 
\cite{shahgir2026vlmsneedwordsvision}. 
At the same time, LLaVA-OneVision shows a more localized pattern, with the largest drops often occurring for \textsc{category-critical tokens}, especially for the captions. These tokens - including nouns, verbs and/or prepositions - correspond to the lexical material that encodes the specific spatial, causal, or temporal contrast between the two candidate answers. This suggests that 
multimodal enrichment is especially concentrated on the task-critical spans that determine which option is visually correct, rather than being distributed more generally across broad linguistic categories.\\
Finally, the Temporal subset again differs from the patterns observed for Spatial and Causal data. First, the use of visual information by linguistic elements within the options appears more irregular, with frequent peaks, drops, and recoveries across layers. Second, compared to the Spatial and Causal subsets, verbs show a stronger dependence on the visual input (Figure \ref{fig:Temporal_ling}). This difference aligns with the nature of the Temporal task: while Spatial and Causal items can often be solved by grounding entities or salient visual states, Temporal items require the model to track actions, event dynamics, and their ordering across the video. The stronger involvement of verbs may therefore reflect the need to process visually grounded event information, rather than only visually grounded entities.\\[4pt]
As an additional robustness check (i.e., to further assess whether these patterns are tied to the original Italian data), we replicated the analysis on English translations of 200-item samples drawn from each of the original MAIA subsets. The results, reported in Appendix~\ref{app:english_masking}, broadly confirm the trends observed in Italian for both region-based and token-level attention knockout. However, although further analysis is needed, the Temporal subset shows partially different dynamics in English, appearing more stable across layers while still differing from Spatial and Causal cases and suggesting that temporal-relation processing may also be affected by the language of the input. 




\begin{table}[t]
\centering
\scriptsize
\setlength{\tabcolsep}{3pt}

\resizebox{\columnwidth}{!}{%
\begin{tabular}{l r *{2}{cc}}
\toprule
\multirow{2}{*}{Temporal marker} &
\multirow{2}{*}{$N$} &
\multicolumn{2}{c}{\textit{Text-only baseline}} &
\multicolumn{2}{c}{\textit{Vision (None)}} \\
\cmidrule(lr){3-4}
\cmidrule(lr){5-6}
& & Qwen & LLaVA & Qwen & LLaVA \\
\midrule
After             & 443 & 0.42 & 0.36 & 0.56 & 0.65 \\
Before            & 115 & 0.82 & 0.75 & 0.70 & 0.40 \\
Beginning / Start & 42  & 0.45 & 0.88 & 0.60 & 0.86 \\
End               & 106 & 0.59 & 0.35 & 0.71 & 0.62 \\
When              & 195 & 0.42 & 0.27 & 0.41 & 0.28 \\
While             & 523 & 0.12 & 0.25 & 0.13 & 0.17 \\
Others            & 176 & 0.63 & 0.60 & 0.66 & 0.68 \\
\bottomrule
\end{tabular}%
}

\caption{Last-layer accuracy by temporal marker. 
}
\label{tab:temporal_markers}
\end{table}

\subsection{The Temporality Issue}
\label{subsec:temporality_issue}
These observations raise two related questions: (i) why is Temporal information processed differently, leading to a different and weaker ability to solve the task?; (ii) why does the Temporal subset not behave similarly to the Causal subset, even though temporal structure is also relevant for identifying cause-effect relations? A possible explanation is that Temporal and Causal items involve different forms of temporal information. In Causal examples, identifying the cause and the effect of an event may often rely on commonsense knowledge (some of which is also available to the language model, and need not rely on the sample-specific visual inputs) and on the recognition of salient visual evidence within individual frames linked to textual representation (e.g., Nouns and Verbs). By contrast, Temporal examples require the model to identify and preserve the sequential order of events, such as whether event A occurs before event B, and in some cases to ground the answer in the absolute temporal position of the video, such as the beginning/ end of the scene. 
The Spatial subset further supports this hypothesis. Although some Spatial examples contain temporal expressions, such as \textit{at the beginning of the scene}, temporality is not the source of the semantic inconsistency between caption and foil. The model can therefore solve the task by retrieving the relevant visual information from a specific moment or frame of the video. In Temporal examples, instead, the temporal relation itself is the critical information that must be disentangled. We therefore hypothesize that the difficulty observed here is caused by the need to recover and maintain sequential information across the video, a process that appears to be particularly challenging for both models, and an issue that recent work has shown to be challenging also for state of the art VLMs~\cite{song-etal-2025-burn,walat_study_2025}.

\paragraph{Temporal markers and linguistic biases}

While this visual-sequential difficulty likely plays a central role, it may not fully explain the below-chance Temporal performance (see Table~\ref{tab:behavioral_accuracy}). To assess whether the linguistic formulation of the candidates also contributes, we identify the temporal markers present in the caption--foil pairs and analyze model preferences for each marker (Table~\ref{tab:temporal_markers}). The results reveal clear construction-specific asymmetries: in the \textit{text-only} condition, both models strongly favor captions containing \textit{Before} over those containing \textit{After} (Qwen: 0.82 vs. 0.42; LLaVA: 0.75 vs. 0.36). Moreover, \textit{When} and \textit{While} remain below chance even in the \textit{none} condition, where full visual input is available (\textit{When}: 0.41/0.28; \textit{While}: 0.13/0.17 for Qwen/LLaVA). With \textit{While} being the largest group, these patterns suggest that 
linguistic biases may contribute substantially to the aggregate below-chance performance, compounding the difficulty of recovering temporal relations from video.


\section{Conclusion}
\label{sec:concl}

In this work, we investigated the internal cross-modal dynamics through which VLMs use visual evidence in a video-based decision-making setting. Using layer-wise Attention Knockout, we traced \textit{video-to-text} information flow by selectively blocking attention pathways between visual tokens and different textual regions of the prompt, focusing on spatial, causal, and temporal reasoning. The analysis is useful to provide insight into where and how visual information enters the textual representations used for model decisions. Overall, our results show that: (i) visual information mainly contributes when the model processes the candidate answer options, rather than the task instruction or the assistant response, suggesting that candidate options are the primary site of cross-modal grounding and that models rely on these visually enriched representations, rather than on the video directly, when selecting the correct candidate; (ii) 
the token-level analysis further clarifies which linguistic elements mediate this process, with nouns acting as semantic anchors for visually grounded entities, verbs becoming more prominent in temporal examples, and, in some cases, the affected words being exactly those that distinguish the correct option from the incorrect one; and (iii) spatial and causal relations follow similar processing dynamics, whereas temporal relations exhibit more fragile behavior, suggesting that models struggle to maintain and exploit sequential information across the video, with this difficulty potentially reflecting, in part, linguistic biases associated with expressions used for describing temporal relations between events within a scene.


\section*{Limitations}

One limitation of this work is that we evaluate only two relatively small VLMs. This restricts the generalizability of our findings across the rapidly evolving landscape of multimodal models. However, our goal was not to provide a leaderboard-style comparison of model performance, but to conduct a controlled diagnostic analysis of how visual information contributes to model predictions. 
This choice allowed us to examine whether similar information-flow patterns emerge across quite different architectures
, while keeping the analysis and visualization of layer-wise interventions manageable within the space constraints of the main paper.\\
A second limitation concerns the scoring formulation used in the mechanistic analyses. While we report both A/B-label scoring and full-sentence scoring as behavioral reference points, we do not systematically analyze the differences between them. Our mechanistic analysis required sentence-level scoring, since we needed access to the token-level structure of each candidate answer and to measure how the plausibility of the full option changes under attention interventions, rather than only whether the model assigns higher probability to the label \textit{A} or \textit{B}. However, the drop observed for Temporal items suggests that the choice of scoring a full sentence may affect model behavior, and should be investigated more directly in future work.

\section*{Acknowledgments}

This work has been carried out while Davide Testa was
enrolled in the Italian National Doctorate on Artificial
Intelligence run by Sapienza University of Rome in collaboration with Fondazione Bruno Kessler (FBK).
HMW is funded by the Dutch Research Council (NWO) through the AiNed Fellowship Grant NGF.1607.22.002. Bernardo Magnini was supported by the PNRR MUR project \href{https://fondazione-fair.it/}{PE0000013-FAIR} (Spoke 2). Albert Gatt was supported by the NWO through the AiNed grant NGF.1609.23.020.

\bibliography{custom}

\appendix

\section{Additional Experiments and Analysis}
\subsection{Text-to-vision Masking Direction}
\label{app:text-to-vision}

In this appendix, we report the complementary analysis conducted with the \textit{text-to-vision} masking direction. While the main paper focuses on the \textit{vision-to-text} setting (Section \ref{sec:exp}), where textual query tokens are prevented from attending to visual tokens, this additional experiment investigates the opposite direction of information flow. Specifically, we test whether visual-token representations are affected when their attention to textual regions is selectively blocked.

This analysis requires a different prompt configuration from the one used in the main experiments. In the standard Video-Text (VT) prompt, the video input precedes the textual prompt, as shown in Figure~\ref{fig:prompt}. Under the causal attention mechanism of decoder-only transformer backbones, each token can only attend to previous tokens in the sequence. As a consequence, if visual tokens appear before textual tokens, they cannot attend to the text, making a \textit{text-to-vision} intervention ill-defined: there would be no text-to-vision attention connections to remove.

\begin{figure}[t]
  \centering
    \includegraphics[width=\linewidth]{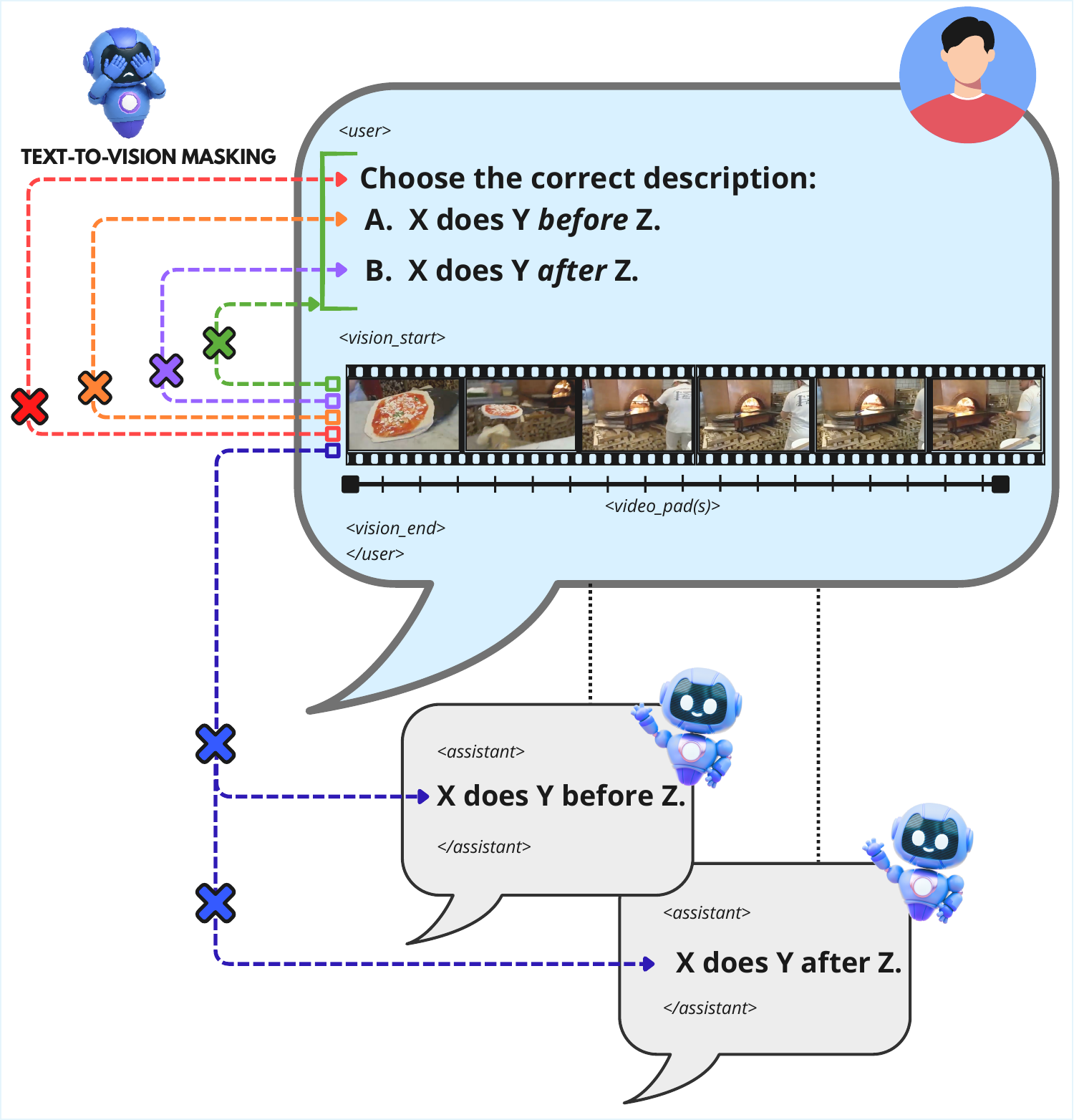}
  \hfill

  \caption{Overview of the Attention Knockout approach with inverted Text-Video prompt used for the \textit{text-to-vision} masking direction.}
  \label{fig:prompt_TV}
\end{figure}

To make this intervention meaningful, we therefore adopted an inverted Text-Video (TV) prompt, where the textual prompt is placed before the video input, as shown in Figure~\ref{fig:prompt_TV}. In this configuration, visual tokens occur after the textual tokens and can therefore attend to them under the standard causal mask. This allows us to selectively block attention from visual query tokens to specific textual key regions, such as the task instruction, option \textit{A}, option \textit{B}, or the full user text. In this way, the \textit{text-to-vision} setup provides a complementary diagnostic view of whether textual information contributes to the construction of visual-token representations.

\begin{figure*}[t]
  \centering
    \includegraphics[width=\linewidth]{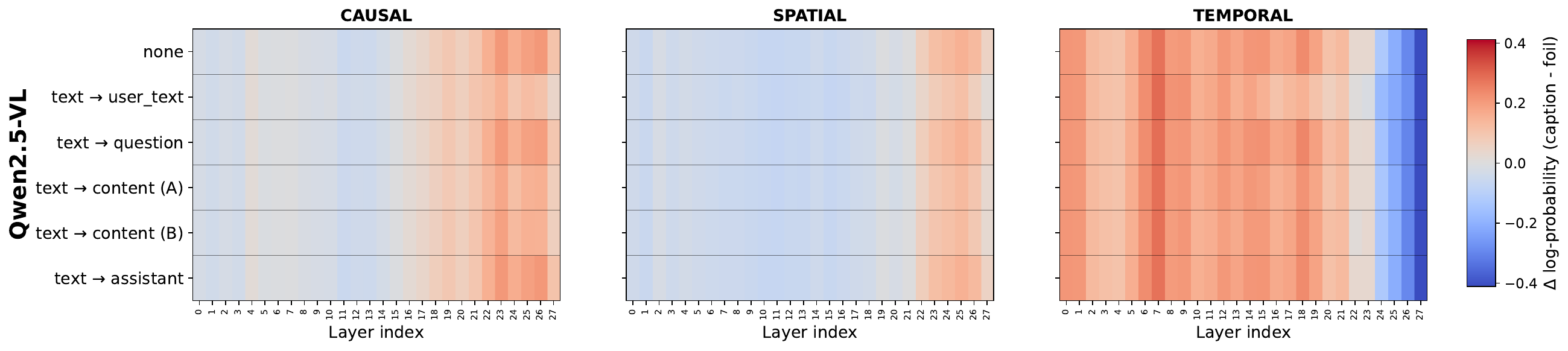}

  \caption{Region-based attention knockout results. Rows indicate the visual prompt region prevented from attending to textual tokens (i.e., \textit{Text} $\rightarrow$ \textit{Region}). Colors show the layer-wise caption-foil log-probability difference.}
  \label{fig:heatmapsTV}
\end{figure*}

We report these results only for Qwen2.5-VL. LLaVA-OneVision is not included in this analysis because its chat template enforces a fixed video-text ordering: even when the input is provided in Text-Video order, the internal template reconstructs the prompt in the default Video-Text configuration. As a result, the inverted prompt required for a valid \textit{text-to-vision} intervention cannot be reliably applied to this model family. Figure~\ref{fig:heatmapsTV} shows the results of the \textit{text-to-vision} region-based knockout for Qwen2.5-VL. Despite the non-standard prompt order, the observed trends are broadly consistent with the main \textit{vision-to-text} analysis discussed in Section~\ref{sec:resDisc}. Blocking attention from visual tokens to textual regions affects the final caption-foil log-probability difference across layers, suggesting that textual regions also contribute to the construction of multimodal representations. However, the magnitude of these effects is generally lower than in the main \textit{vision-to-text} setting.

The \textit{text-to-vision} results show a weaker and less localized pattern than the main \textit{vision-to-text} analysis. The most visible effects emerge when masking the broader \textit{user text} region, whereas the isolated masking of option \textit{A} and option \textit{B} produces only marginal changes. This does not mean that the candidate descriptions are irrelevant, since both options are contained within the broader \textit{user text} span. Instead, it suggests that in the inverted Text-Video configuration, the contribution of textual information to visual-token representations is detectable mainly at the level of the full textual context, rather than at the level of the individual answer-specific regions.

Accordingly, we treat the \textit{text-to-vision} analysis as a complementary control rather than as a fully symmetrical counterpart to the main \textit{vision-to-text} experiments. The results do not contradict the main findings, but they indicate that the reverse direction produces weaker and less clearly localized effects. This pattern may also be influenced by the non-standard nature of the Text-Video prompt order, which is used here only to make the intervention well-defined under causal attention.

Finally, the Temporal subset shows a behavior that differs from the Spatial and Causal subsets, in line with the main results. This reinforces the interpretation that temporal reasoning poses a qualitatively different challenge for the model, and that its difficulty is not simply an artifact of the standard Video-Text prompt order. Overall, the \textit{text-to-vision} analysis supports the conclusions drawn in the main paper while remaining complementary to them: the main evidence points to a dominant \textit{video-to-textual-representation} pathway, whereas this appendix shows that the reverse conditioning direction produces compatible, but weaker, effects.

\begin{figure*}[t]
  \centering
    \includegraphics[width=\linewidth]{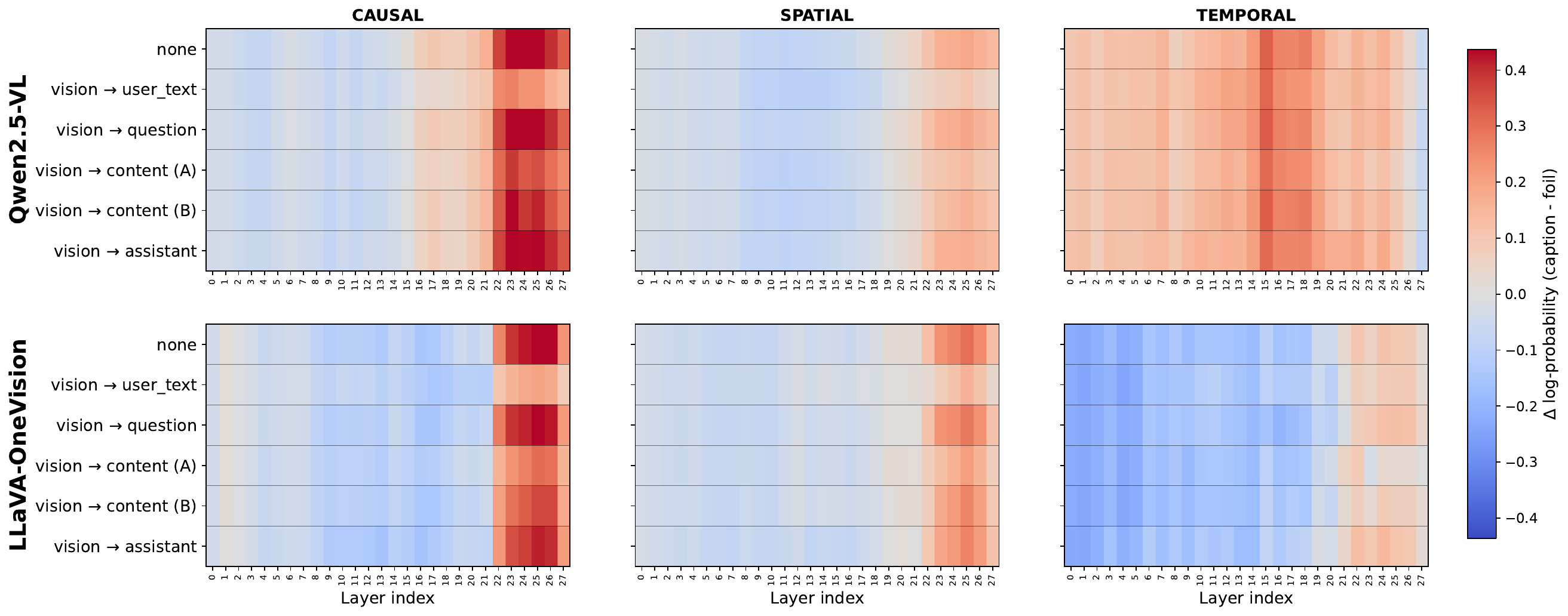}

  \caption{Region-based attention knockout results for English translations of MAIA subdatasets. Rows indicate the textual prompt region prevented from attending to visual tokens (i.e., \textit{Vision} $\rightarrow$ \textit{Region}). Colors show the layer-wise caption-foil log-probability difference.}
  \label{fig:heatmaps_eng}
\end{figure*}

\begin{figure}[t]
  \centering    \includegraphics[width=\linewidth]{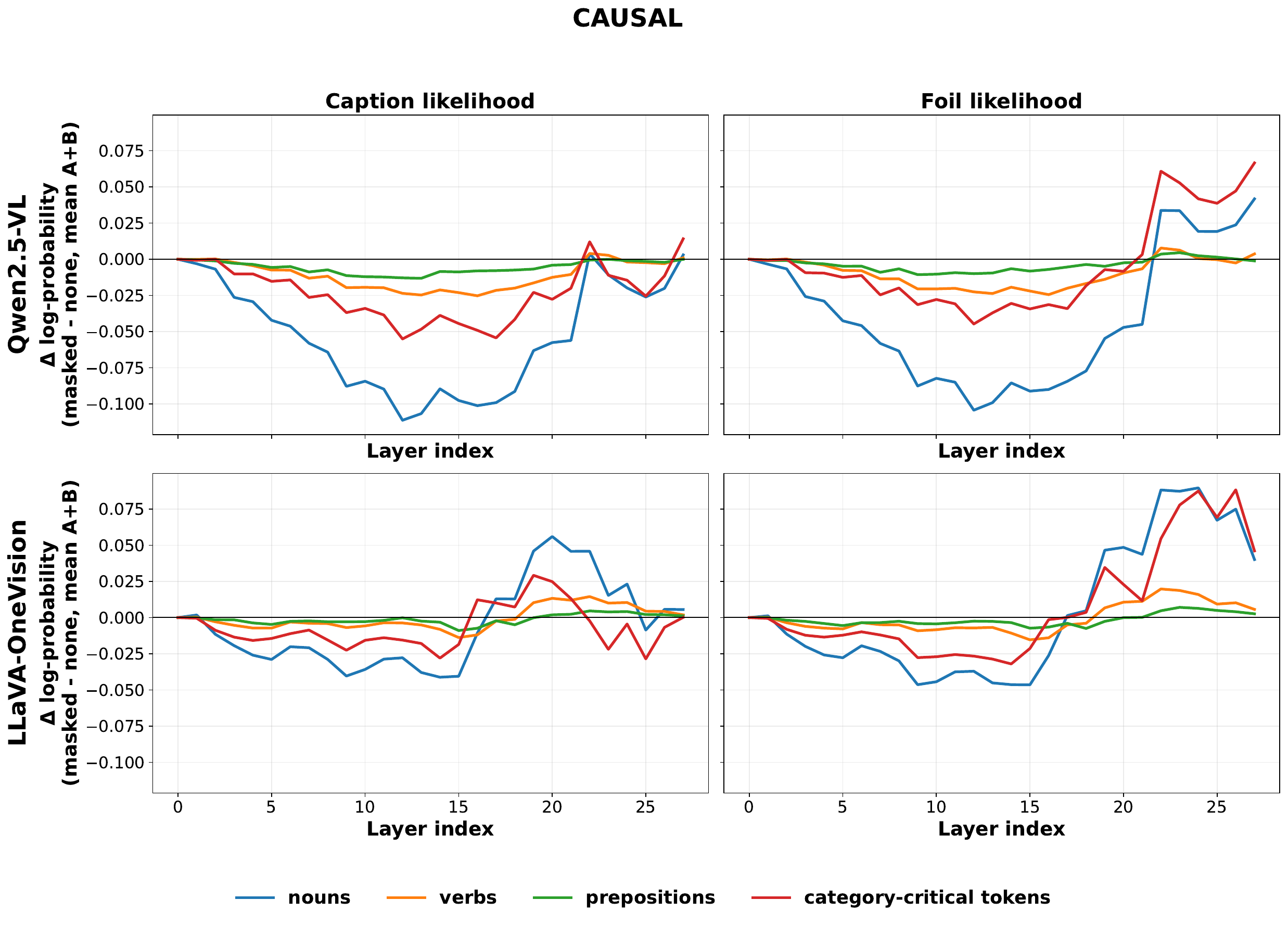}

  \caption{Layer-wise effect of token-level attention knockout on (causal) candidate log-probability. English data}
  \label{fig:Causal_ling_eng}
\end{figure}

\begin{figure}[t]
  \centering
    \includegraphics[width=\linewidth]{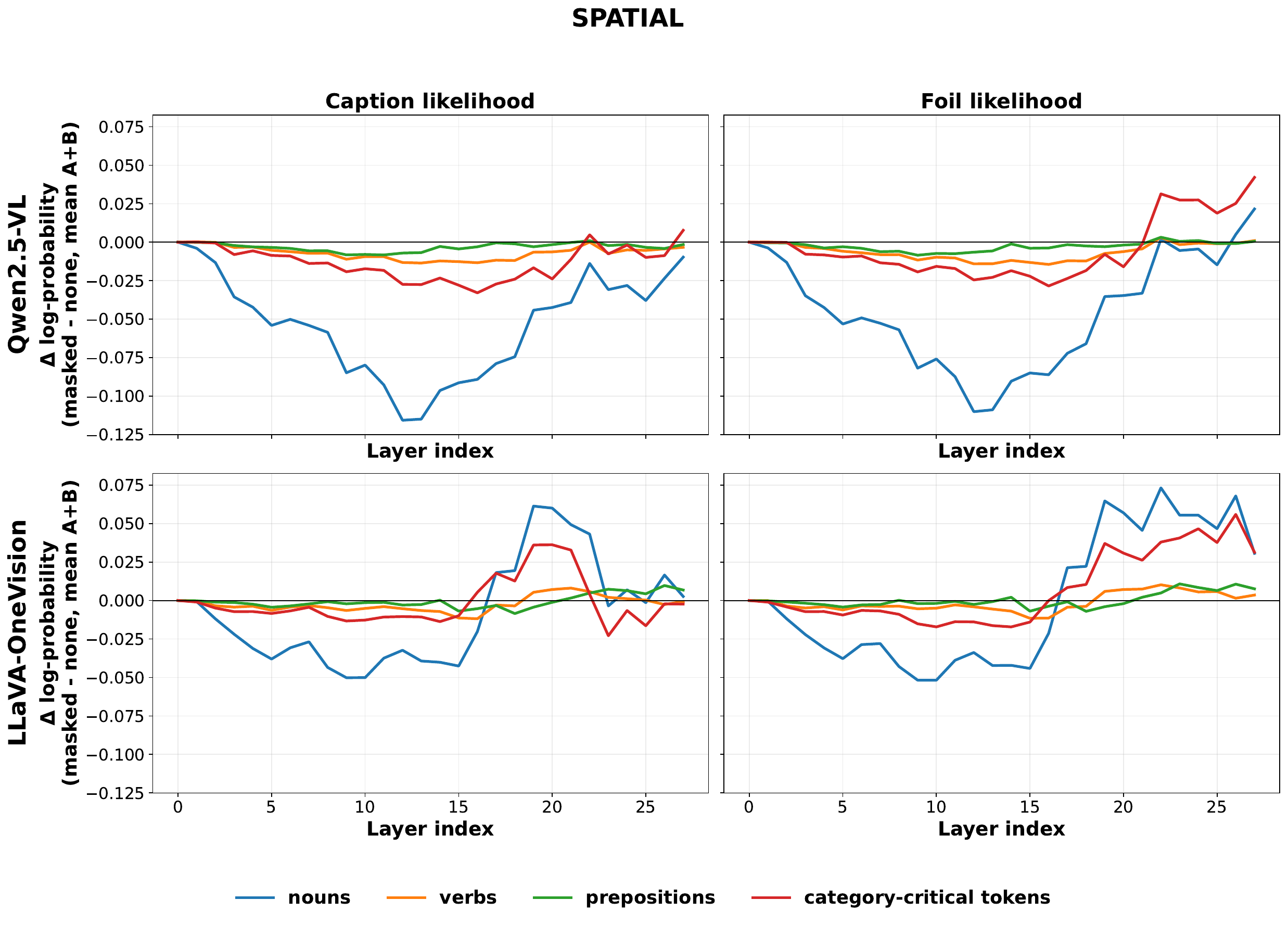}

  \caption{Layer-wise effect of token-level attention knockout on (spatial) candidate log-probability. English data.}
  \label{fig:Spatial_ling_eng}
\end{figure}

\begin{figure}[t]
  \centering
    \includegraphics[width=\linewidth]{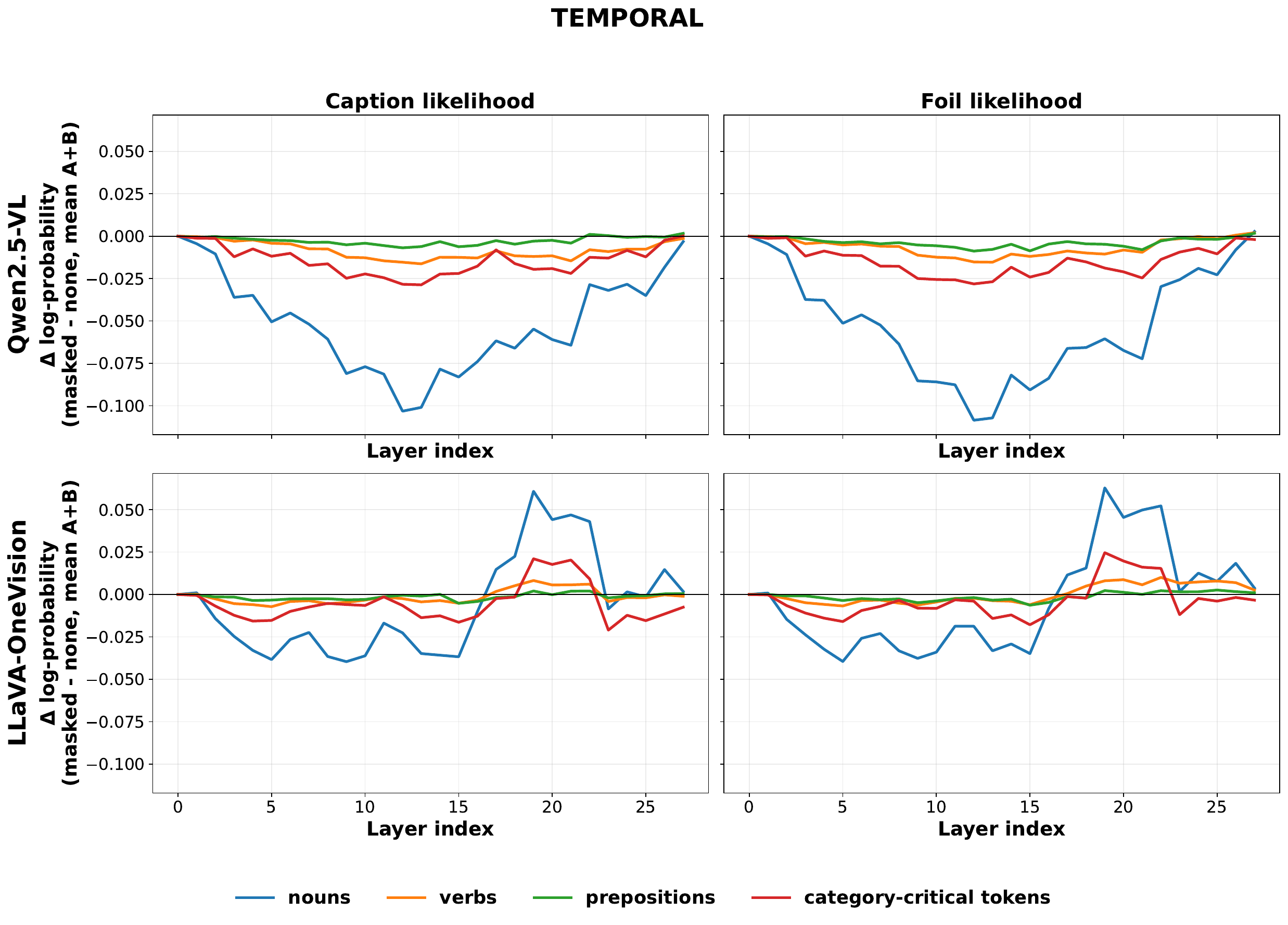}
  
  \caption{Layer-wise effect of token-level attention knockout on (temporal) candidate log-probability. English data.}
  \label{fig:Temporal_ling_eng}
\end{figure}

\begin{table*}[t]
  \centering
  \small
  \begin{tabular}{lll}
    \hline
    \textbf{Dataset} & \textbf{Option type} & \textbf{Sentence} \\
    \hline

    \multirow{2}{*}{\textsc{Spatial}}
      & \textit{Caption} & Alla fine della scena, la bambina sta camminando \textbf{su un tappeto}. \\
      & \textit{Foil}    & Alla fine della scena, la bambina sta camminando \textbf{per strada}. \\

    \multirow{2}{*}{\textsc{Temporal}}
      & \textit{Caption} & \textbf{Subito dopo aver tagliato} la pizza, la ragazza la porge alla persona che ha di fronte. \\
      & \textit{Foil}    & \textbf{Prima di tagliare} la pizza, la ragazza la porge alla persona che ha di fronte. \\

    \multirow{2}{*}{\textsc{Causal}}
      & \textit{Caption} & La ragazza è infastidita a causa di  \textbf{uno scarafaggio}. \\
      & \textit{Foil}    & La ragazza è infastidita a causa di  \textbf{un forte rumore}. \\

    \hline
  \end{tabular}
  \caption{Original Italian examples of caption-foil pairs from the three MAIA subsets.}
  \label{tab:maia_examples_ita}
\end{table*}

\subsection{Cross-lingual analysis}
\label{app:english_masking}

To test whether the observed information-flow patterns depend on the original Italian formulation of the MAIA data, we repeated the same attention knockout experiments on English translations of 200-item samples drawn from each of the three subsets. This analysis is particularly relevant because both tested models are likely to have been exposed more extensively to English during training, and may therefore process English prompts more robustly than Italian ones.

Figure~\ref{fig:heatmaps_eng}, reports the region-based results, while Figures~\ref{fig:Causal_ling_eng}, \ref{fig:Spatial_ling_eng}, and \ref{fig:Temporal_ling_eng} report the token-level results. Overall, the English setting confirms the main trends observed on the Italian data. In the region-based analysis, the strongest effects still emerge when visual access is blocked for textual regions related to the candidate options, supporting the interpretation that these regions act as the main sites where cross-modal information is integrated. In the token-level analysis, visually grounded lexical categories show patterns comparable to those observed in the original Italian setting, with nouns remaining central for grounding. However, the English Temporal subset exhibits partially different dynamics. In the region-based analysis, the layer-wise behavior appears more stable than in the Italian setting, although the two models still show opposite tendencies in their final predictions. Moreover, token-level interventions suggest a reduced impact of verbs compared to the original Italian data. While these observations are not sufficient to draw definitive conclusions, they indicate that the internal processing of temporal relations may also be influenced by the linguistic formulation of the input, opening interesting directions for future cross-lingual analyses.

\subsection{Correct Vs. Incorrect Split Analysis}
\label{app:split_analysis}

\begin{figure}[t]
  \centering
    \includegraphics[width=\linewidth]{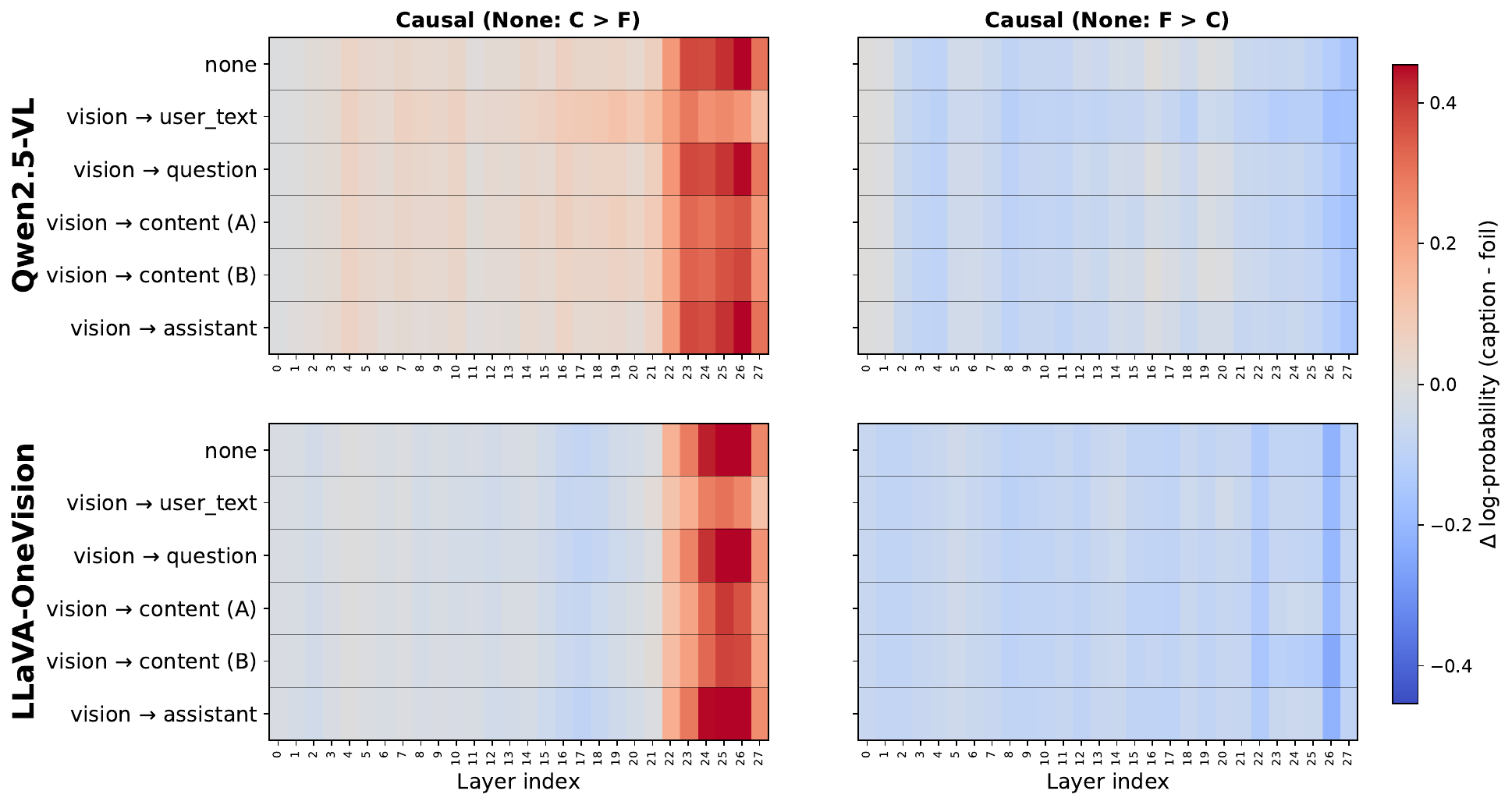}
  \caption{$\Delta$ log-probability (caption $-$ foil) across layers and token regions  on causal items, split by model outcome: correct predictions ($C > F$, left) and incorrect predictions ($F > C$, right).}
  \label{fig:split_causal}
\end{figure}

\begin{figure}[t]
  \centering
    \includegraphics[width=\linewidth]{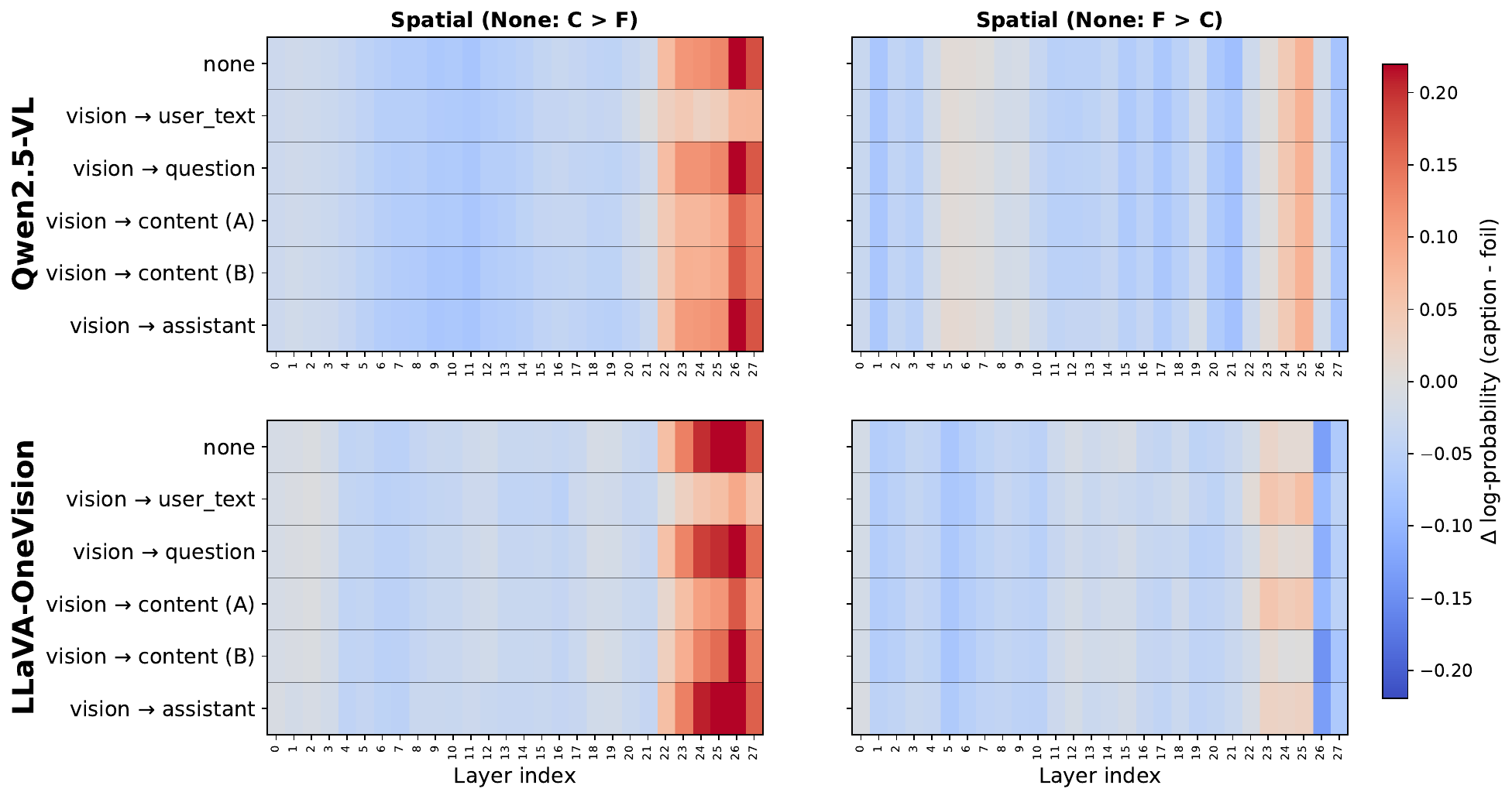}
  \caption{$\Delta$ log-probability (caption $-$ foil) across layers and token regions  on spatial items, split by model outcome: correct predictions ($C > F$, left) and incorrect predictions ($F > C$, right).}
  \label{fig:split_spatial}
\end{figure}

\begin{figure}[t]
  \centering
    \includegraphics[width=\linewidth]{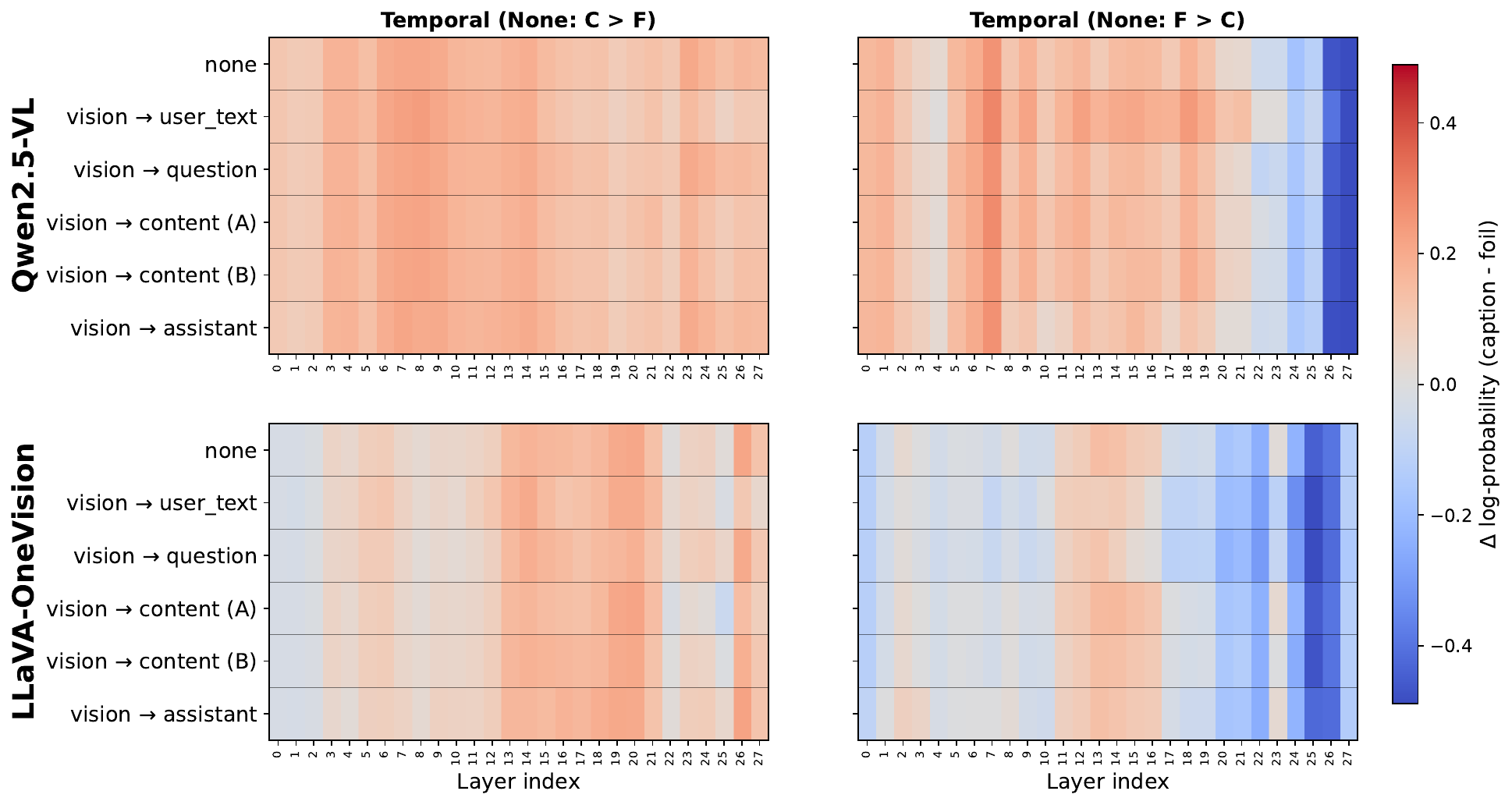}
  \caption{$\Delta$ log-probability (caption $-$ foil) across layers and token regions  on temporal items, split by model outcome: correct predictions ($C > F$, left) and incorrect predictions ($F > C$, right).}
  \label{fig:split_temporal}
\end{figure}

In the main analysis, we report heatmaps averaged over all items, independently of whether the model assigns higher probability to the caption or to the foil in the final layer. To better understand whether the observed layer-wise patterns are driven by correct or incorrect decisions, we further split the results according to the final caption--foil preference in the \textit{none} condition, where no attention masking is applied. Specifically, we distinguish cases in which the final score satisfies $C>F$, corresponding to a correct preference for the caption, from cases in which $F>C$, where the model assigns higher probability to the foil. Figures~\ref{fig:split_causal}, \ref{fig:split_spatial}, and \ref{fig:split_temporal} report the corresponding heatmaps for the Causal, Spatial, and Temporal subsets, respectively. Overall, the split analysis shows that the averaged patterns observed in the main paper are stable for Causal and partially for Spatial items. 

For Causal items, the distinction between correct and incorrect cases is highly consistent across both models. When the model is correct ($C>F$), the caption becomes increasingly favored in the final layers. Conversely, when the model is incorrect ($F>C$), the foil remains preferred in the final layers, with a relatively stable negative pattern. This suggests that, for Causal items, the final outcome is largely reflected in a coherent layer-wise separation between correct and incorrect trajectories.

Spatial items show a less clear pattern. In the $C>F$ split, both models converge toward the caption in the final layers, similarly to the Causal subset. However, in the $F>C$ split, the trajectory is less stable. For Qwen2.5-VL, the model appears to favor the caption around layers $22$--$25$ before shifting toward the foil at the final layer. LLaVA-OneVision shows a similar trend, with a gradual movement toward the caption starting around the same layers, followed by a final shift toward the foil. Thus, although Spatial items are more similar to Causal than to Temporal in the averaged analysis (Figure~\ref{fig:heatmaps}), the split view reveals that incorrect Spatial decisions can also be resolved very late in the network, suggesting uncertainty between the caption and the foil.

The Temporal subset shows the most distinctive behavior. In the averaged heatmaps reported in the main paper, Temporal items appear to follow an opposite trajectory compared to Spatial and Causal items, with the foil receiving higher probability in the later layers. However, the split analysis in Figure~\ref{fig:split_temporal} reveals that this pattern is mainly driven by the $F>C$ cases, i.e., by the items for which the model ultimately assigns higher probability to the foil in the \textit{none} condition. When the model is correct ($C>F$), the deltas are positive from the middle layers in both models, and even from earlier layers in Qwen2.5-VL. Although both models show some uncertainty in the penultimate layers, approximately between layers $22$ and $25$, they ultimately converge toward higher probability for the caption by layer $26$. In contrast, when the model is incorrect ($F>C$), the foil begins to receive higher probability only in the late layers: around layer $22$ for Qwen2.5-VL, and somewhat earlier for LLaVA-OneVision, which nevertheless shows greater variability between the two alternatives. These results indicate that the apparent Temporal inversion observed in the averaged heatmaps should not be interpreted as a uniform property of all Temporal items. Rather, it reflects a late-layer shift specific to the incorrect subset. In other words, for Temporal items, the choice between caption and foil appears to be determined late in the network: in successful cases, the model eventually stabilizes on the caption, whereas in unsuccessful cases the foil becomes dominant only in the final layers. This supports the interpretation that Temporal reasoning is more fragile and depends on the model's ability to preserve and exploit sequential information across the video until the last stages of computation.

\section{Additional Materials}
\label{app:add_data}

This Appendix section presents additional materials not included in the main paper.\\
Table \ref{tab:maia_examples_ita} shows the original Italian caption-foil pairs examples which were presented in English in Section~\ref{subsec:data}. 

\section{Experiments}
\label{sec:appendixB}
This appendix section will contain additional details on our experimental settings, including a description of the VLMs used. Attention Knockout experiments of subsets of MAIA were conducted using A$100$ GPUs ($40$GB). Overall, the total computational budget was on the order of $\sim$2,000 GPU hours.

\subsection{Models tested}
\label{sec:models_app}

\begin{description}
[style=unboxed,leftmargin=0cm,noitemsep]
    \item[LLaVa-OneVision.] \cite{li2024llava1V, llavanextvideo}: $7$B parameter model that builds on the LLaVA framework with a Qwen2 LLM backbone to serve as a general-purpose vision-language assistant; pre-trained on extensive multimodal data to deliver robust cross-modal reasoning. \textit{Hugging Face} model: lmms-lab/llava-onevision-qwen2-7b-ov.
    \item[Qwen2.5-VL.] \cite{qwen2.5technicalreport}: $7$B parameter VLM of the Qwen family using the \textit{Qwen}$2.5$ LLM decoder; key enhancements are related to grounding, working with longer videos and capturing events. It was pre-trained on comprehensive visual and textual datasets and fine-tuned for detailed, context-aware responses. \textit{Hugging Face} model: Qwen/Qwen2.5-VL-7B-Instruct.
\end{description}

\subsection{Declaration on Generative AI}

 During the preparation of this work, the author(s) used ChatGPT in order to: Grammar and spelling check and help in coding. After using these tool, the authors reviewed and edited the content as needed and take full responsibility for the publication’s content. 
 
\end{document}